\documentclass[runningheads]{llncs}
\usepackage{amsmath}
\usepackage{amssymb}
\usepackage{mathtools}
\usepackage{stackrel}
\usepackage{algorithmic}
\usepackage{array}
\usepackage{tikz}
\usetikzlibrary{bayesnet}
\usepackage{hyperref}
\usepackage{scalerel,stackengine}
\usepackage{colortbl}
\usepackage{graphicx}
\usepackage{dirtytalk}
\usepackage{float}
\usepackage{etoolbox}

\newtoggle{full}
\toggletrue{full}

\iftoggle{full} {
\newcommand{\scaleFactor}{1}
}{ 
\newcommand{\scaleFactor}{1.2}
} 

\newcommand{\bigSkipOnNonfull}{
\iftoggle{full} {
}{ 
\bigskip
} 
}

\newcommand{\Ln}[1] {\mathop{\mathrm{ln}}\left( #1 \right)}

\newcommand{\Rank}[1] {\mathop{\mathrm{rank}}\left( #1 \right)}

\renewcommand{\Vec}[1] {\underline{#1}}

\newcommand{\Diag}[1] {\mathop{\mathrm{diag}}\left( #1 \right)}

\newcommand{\Ignore}[1] {}

\newcommand{\GroupVKEps} {\Group{V}_k^{\epsilon}}

\newcommand{\Density}[1]{\ensuremath{f\left(#1\right)}}
\newcommand{\DensityHat}[1]{\ensuremath{\hat{f}\left(#1\right)}}

\newcommand{\DensityNML}[1]{\ensuremath{f^{\mathrm{NML}}}\left(#1\right)}

\newcommand{\Conditional}{\;|\;}
\newcommand{\Given}{\;;\;}

\newcommand{\Probability}[1]{P\left(#1\right)}
\newcommand{\Kullback}[2]{D\left({#1} \Conditional {#2}\right)}
\newcommand{\Sin}[1]{\mathrm{sin}\left(#1\right)}
\newcommand{\Cos}[1]{\mathrm{cos}\left(#1\right)}

\newcommand{\Group}[1] {\mathcal{#1}}
\newcommand{\Set}[1] {\left\{ #1 \right\}}
\newcommand{\SizeOf}[1] {\left\lvert #1 \right\rvert}

\newcommand{\EqnIndent} {\;\;}
\newcommand{\EqnNL}[1] {\\ &\EqnIndent #1}

\newcommand{\NormalD}{\ensuremath{\mathcal{N}}}
\newcommand{\UniformD}{\ensuremath{\mathcal{U}}}

\newcommand{\Frobenius}[1]{\ensuremath{\left\lVert #1 \right\rVert_F}}
\newcommand{\FrobeniusSq}[1]{\ensuremath{\left\lVert #1 \right\rVert_F^2}}

\newcommand{\LTwo}[1]{\ensuremath{\left\lVert #1 \right\rVert}}

\newcommand \Dd[1]  { \,\textrm d{#1}                       }   
\newcommand {\IntStart}[2] {\int\limits_{#1}^{#2}}
\newcommand {\IntEnd}[1] {\Dd{#1}}
\newcommand \Int[4]{ \IntStart{#1}{#2}{#3}\IntEnd{#4} }   

\newcommand \OptFor[2] {\ensuremath{\hat{#1}\left(#2\right)}}

\newcommand \LetteredLEq[1] {\mathrel{\overset{\makebox[0pt]{\mbox{\normalfont\tiny\sffamily (#1)}}}{\leq}}}
\newcommand \LetteredGEq[1] {\mathrel{\overset{\makebox[0pt]{\mbox{\normalfont\tiny\sffamily (#1)}}}{\geq}}}
\newcommand \LetteredEq[1] {\mathrel{\overset{\makebox[0pt]{\mbox{\normalfont\tiny\sffamily (#1)}}}{=}}}
\newcommand \LetteredSimEq[1] {\mathrel{\overset{\makebox[0pt]{\mbox{\normalfont\tiny\sffamily (#1)}}}{\simeq}}}

\begin{document}
\title{Determining Principal Component Cardinality\\through the\\Principle of Minimum Description Length }
\titlerunning{Principal Components through NML}
%
\author{Ami Tavory\inst{1}\orcidID{0000-0001-5981-1519}}
\authorrunning{A.\ Tavory}
%
\institute{Facebook Research, Core Data Science\\
\email{atavory@fb.com}}
\maketitle              
\begin{abstract}
PCA (Principal Component Analysis) and its variants are ubiquitous techniques for
matrix dimension reduction and reduced-dimension latent-factor extraction.
One significant challenge in using PCA, is the choice of the number of principal components.
The information-theoretic MDL (Minimum Description Length) principle gives objective compression-based criteria for model selection, but
it is difficult to analytically apply its modern definition - NML (Normalized Maximum Likelihood) - to the problem of PCA.
This work shows a general reduction of NML problems to lower-dimension problems. Applying this reduction, it bounds the NML of PCA, by terms of the NML of linear regression, which are known.

\end{abstract}
\section{Introduction}\label{sec:Intro}

\subsection{The Problem of Principle Component Dimension Selection}\label{subsec:IntroPCAProb}

Let $X$ be an an $n \times m$ matrix. In machine learning, it is very common to approximate it by a ``simpler'' product of matrices $W$ and $Z^T$ of lower dimensions $n \times k$ and $k \times m$, respectively (for $k \lneq m$). Among others, these include Probabilistic Principal Component Analysis, Independent-Factor Analysis, and Non-Negative Matrix Factorization (see \cite{Hastie01ElementsStatisticalLearning,Udell2016GLRM,BOKDE2015136}). We will focus specifically on the simple PCA (Principal Component Analysis),
\begin{equation}\label{eqn:MainMinObjective}
\arg \min_{W, Z:\; \Rank{W} = \Rank{Z} = k} \FrobeniusSq{X - W Z^T}
.
\end{equation}

\bigSkipOnNonfull
The lower-dimension product is not guaranteed to losslessly approximate the original matrix. In fact, the famous Eckart-Young-Mirsky Theorem -  whose properties we will use throughout - essentially guarantees some loss:
\begin{theorem}\label{thm:SVD}
\textbf{(Eckart-Young-Mirsky)}
Let $X = U \Lambda V^T$ be the SVD (singular value decomposition) of $X$, with $\Lambda = \Diag{\lambda_1, \ldots, \lambda_m}$, and $U$ and $V$ unitary. Let $U_k$ and $V_k$ be the matrices of the first $k$ columns of $U$ and $V$, respectively. Then
\begin{equation}\label{eqn:SVDEckartYoungMirsky}
\begin{split}
& \FrobeniusSq{X - W Z^T}
\geq
\FrobeniusSq{X - U_k \Diag{\lambda_1, \ldots, \lambda_k} V_k}
=
\sum_{i = k + 1}^m \left[ \lambda_i^2 \right]
,
\end{split}
\end{equation}
and so
$W = U_k \Diag{\lambda_1, \ldots\lambda_k}$, $Z = V_k$, is optimal.
\end{theorem}

\bigSkipOnNonfull
The motivation for the reduced dimension, is uncovering a structure that is, in some sense, ``truer'', or ``more useful''. To quote \cite{Jolliffe1986PCA}:

\say{
The central idea of principal component analysis is to reduce the dimensionality of a data set in which there are a large number of interrelated
variables, while retaining as much as possible of the variation present in
the data set. This reduction is achieved by transforming to a new set of
variables, the principal components, which are uncorrelated, and which are
ordered so that the first few retain most of the variation present in all of
the original variables.
}

As the theorem shows, though, loss minimization, in itself, will not lead us to the reduced dimension - it will always favor the maximum number of components.

\subsection{The Principles of MDL and NML}\label{subsec:IntroPrinciplesMDLN
ML}

The MDL (minimum description length) principle (see \cite{Hansen2001Model,MyungNavarroPitt2006ModelSelection,Grunwald05ATutorial,Rissanen1987Stochastic,Rissanen1989Stochastic}) is an information-theoretic method for model selection. Probability-theory approaches to model selection - both frequentist and Bayesian - assume that there exists a true probability distribution from which the observed data were sampled. The goal is to optimize a model subject to this (indirectly-observed) distribution. MDL is similar in philosophy to Occam's Razor (see \cite{Blumer1987Occam}). The goal is to find a model optimizing the total description length of the model and the observed data. There is no assumption that a true probability was approximated, or that it even exists. We will see that avoiding this assumption leads to a form of online optimality.

\bigSkipOnNonfull
How can we objectively quantify a description length? Given a probability distribution, information theory gives an objective code length through entropy \cite{Cover2006Elements}, but assumptions on the probability distribution are precisely what we wish to avoid. In \cite{Rissanen2000StrongOptimality}, Rissanen formulated the question as a minimax problem, namely the smallest regret relative to all possible codes under mild conditions. He showed that the NML (Normalized Maximum Likelihood) (see \cite{Rissanen2000StrongOptimality,Barron1998MDLCodingModeling}) is the solution to this problem.
\begin{definition}\label{def:NML}\textbf{Normalized Maximum Likelihood}
Let $X$ be distributed  by a model specified by some parameter(s) $\Phi$. The NML is defined as
\begin{equation}\label{eqn:NMLDef}
\DensityNML{X}
=
{
  \DensityHat{X \Given \OptFor{\Phi}{X}}
  \over
  \Int{} {} {\DensityHat{Y \Given \OptFor{\Phi}{Y}}} {Y}
},
\end{equation}
where
\begin{itemize}
  \item $\OptFor{\Phi}{X}$ is the maximum likelihood (ML) estimator
of $\Phi$ given $X$.
  \item $\DensityHat{Y \Given \OptFor{\Phi}{Y} }$ is the ML of $Y$
assuming that the true parameters are $\OptFor{\Phi}{Y}$.
\end{itemize}
\end{definition}

The logarithm of the right-hand side of Equation (\ref{eqn:NMLDef}) is the \textit{stochastic complexity}, and the logarithm of its denominator is the \textit{parametric complexity}. It can be shown that choosing between different $\Phi$ based on maximizing (\ref{eqn:NMLDef}), is optimal in a prequential sense (see \cite{Dawid1999Prequential}).

\subsection{Main Contribution: Applying NML to PCA}\label{subsec:IntroPCANML}

Conceptually, it is possible to calculate the NML of PCA, by inserting equation (\ref{eqn:ItIsJustEigenvalues}) into equation (\ref{eqn:NMLDef}). Unfortunately, evaluating the denominator requires integrating over the eigenvalues of arbitrary matrices, which is difficult. Instead, in the rest of this paper, we avoid this by bounding the NML of PCA by reducing it to the NML of linear regression (see \cite{Rissanen1999MDLDenoising}), resulting in the following theorem:
\begin{theorem}\label{thm:BoundPCANML}
Let $s\left(X \Given k\right)$ be the stochastic complexity of a $k$-dimensional PCA reduction of $X$. Then
\begin{equation}
\begin{split}
&
s(X \Given k)
\EqnNL{\simeq}
\left( nm - kn \right) \Ln{\sum_{i = k + 1}\left[\lambda_i^2\right]}
+ nk \Ln{ \FrobeniusSq{ X^T X}}
\EqnNL{+}
(mn - kn - 1) \Ln{ mn \over mn - kn}
-
(nk + 1) \Ln{nk}
+ \Delta s
,
\end{split}
\end{equation}
where
\begin{equation}
\begin{split}
0 &\leq \Delta s \leq mk \Ln{2 \over m \epsilon}
.
\end{split}
\end{equation}
\end{theorem}

This means that the number of dimensions can be chosen, by optimizing the above for $k$.

\subsection{Outline}\label{subsec:IntroOutline}

 We continue this section with definitions and notations, and related work. Section \ref{sec:ChallengeMainIdea} shows the main idea of NML reduction via elimination of some of the optimization parameters. We use this to reduce the problem of PCA NML to linear-regression NML. Section \ref{sec:TheReductions} details the specific reductions.
\iftoggle{full} {
 Section \ref{sec:NumericalExperiments} shows numerical experiments.
}{ 
} 
 Section \ref{sec:Conclusions} concludes and discusses further work.

\subsection{Definitions and Notations}\label{subsec:IntroDefinitionsAndNotations}

We will use lowercase letters ($s$) for scalars, underlined lowercase letters ($\Vec{x}$) for column vectors, uppercase letters ($X$) for matrices, and calligraphic ($\Group{B}$) for sets. A single subscript for a matrix denotes a matrix row ($X_i$). $\Density{x}$, $\Density{x \Given y}$, $\Density{x \Conditional y}$ denote the density of some $x$, the density of some $x$ assuming some other parameter is $y$, and the density of some $x$ conditional on some other random variable being $y$, respectively.
$
\Frobenius{X} = \left( \sum_{i, j} X_{i, j}^2 \right)^{1 \over 2}
$
is the Forbenius norm, and $\Kullback{x}{y}$ is the Kullback-Leibler distance.

\subsection{Related Work}\label{subsec:IntroRelatedWork}

\cite{BOKDE2015136,Udell2016GLRM} contain excellent overviews of matrix
factorization; in particular, PCA appears in the classic \cite{EckartYoungMinsky1936ApproximationOneMatrix}.
\cite{MyungNavarroPitt2006ModelSelection,Hansen2003MDLGLM,Rissanen2000StrongOptimality,Grunwald05ATutorial,Hansen2001Model,Barron1998MDLCodingModeling} describe MDL and NML, in
particular, for model selection. \cite{Rissanen1999MDLDenoising,Rissanen2000StrongOptimality} show closed forms of
linear-regression NML. \cite{Josse2012SelectingNumCompPCAUsingCV} uses cross validation approximations for PCA dimension estimation,  \cite{Choi2017SelectingTheNumber} does so using an
analysis of the conditional distribution of the singular values
of a Wishart matrix, \cite{Hoyle2008Automatic} uses a Bayesian approach, \cite{Zhu2006AutomaticDS} uses patterns in the scree plots, and \cite{Donald1993StoppingRules} compares statistical and heuristic approaches to this problem. To the best of my knowledge, previous works did not apply the modern form of the MDL principle to the problem of PCA dimension selection.

\section{NML reduction via Elimination of Optimization Parameters}\label{sec:ChallengeMainIdea}

\begin{figure}[ht]
\centering
\caption{Equivalent factor graph of PCA. The dimension $k$ is a-priori uniform, and the observed matrix $X$ is the product of the score and loadings matrices, with additive noise $\Upsilon$ distributed i.i.d.\ $ \NormalD\left(0, \tau I_k\right)$.}
\label{fig:PCAOrig}
\scalebox{\scaleFactor}{
\begin{tikzpicture}
    \node[obs] (x) {$X$};
    \node[latent, above=of x, xshift=-1.2cm] (w) {$W_k$};
    \node[latent, above=of x, xshift=1.2cm] (v) {$V_k$};
    \node[latent, above=of v, xshift=-1.2cm] (k) {$k$};
    \node[latent, right=1.8cm of x] (ups) {$\Upsilon$};
    \factor[right=0.5cm of ups] {ne} {above:\NormalD} {} {}; %
    \node[const, right=0.3cm of ne] (tau) {$\tau$};
    \factor[right=0.5cm of k] {kd} {above: \UniformD} {} {}; %

    \edge {kd} {k} ; %
    \edge {tau} {ups} ; %
    \edge {k} {w, v} ; %
    \edge {w, v, ups} {x} ; %

    {\tikzset{plate caption/.append style={above=1pt of #1.south west}}
    \plate {wx} {(w)(x.south east)(w.south west)} {$n$} ;
    }
    \plate {vx} {(v)(x)(wx.north east)(x.south west)} {$m$} ;
    \plate {upsn} {(ne.north)(ups.south west)(ups.north west)} {$nm$} ;
\end{tikzpicture}
}
\end{figure}
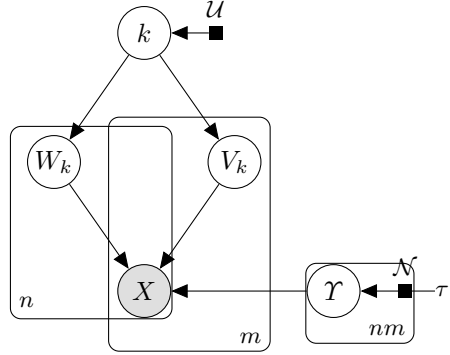

Consider the generative form of (\ref{eqn:MainMinObjective}), shown in
the factor diagram (see \cite{Dietz2010DirFactorNotation}) in Figure \ref{fig:PCAOrig}. In this model, $k \sim \UniformD(1, m)$ determines the dimension of $W_k$ and $V_k$. $X = W_k V_k^T + \Upsilon$, where $\Upsilon \sim \NormalD\left(0, \tau I_k\right)$. Note that they do not appear in the original problem (at least in this form), but the problems are effectively equivalent. The distribution of $k$ hardly affects the stochastic complexity (see \cite{Rissanen1989Stochastic}, Chapter 5), and any distribution assigning a positive probability to any value of $1, \ldots, m$ could be used. Regarding the Gaussian additive noise $\Upsilon$,
\begin{equation*}
\label{eqn:ItIsJustEigenvalues}
\begin{split}
&\arg \max_{W_k, V_k} \Density{X \Given k}
=
\arg \max_{W_k, V_k}
{1 \over \left( 2 \pi \tau \right)^{{nm \over 2}}}
e^{- {\FrobeniusSq{X - W_k V_k^T} \over 2 \tau^2}}
\LetteredEq{a}
\sum_{i = k + 1}^m\left[ \lambda_i^2 \right],
\end{split}
\end{equation*}
where (a) follows from Theorem \ref{thm:SVD}.

\bigSkipOnNonfull
Now consider the generative model in Figure \ref{fig:LinearRegressionNML} (discussed in greater detail in Section \ref{sec:TheReductions}), where both the number of parameters and the loadings matrix are known. This easier problem is more similar to linear regression, whose NML is known (see \cite{Rissanen1999MDLDenoising}). Of course, in the original problem, the loadings matrix is not known, but rather optimized as well. The following Lemma, however, relates the NML of a problem depending on a number of parameters, to the the same problem where one of them is fixed.

\begin{lemma}\label{lem:NMLReplaceOptBySpecific}
Let $\Group{B} = \Set{b_1, \ldots, b_{\ell}}$ be a finite set (for some $\ell$). Then
\begin{equation}
\label{eqn:NMLReplaceOptBySpecificUpper}
\begin{split}
    &\Int{} {} {\DensityHat{X \Conditional \OptFor{A}{X}, \OptFor{b}{X}}} {X}
    \leq
    \sum_{b \in \Group{B}} \Int{} {} {\DensityHat{X \Conditional \OptFor{A}{X}, b}} {X}
    .
\end{split}
\end{equation}

Furthermore, if
\begin{equation}
\label{eqn:NMLReplaceOptBySpecificLowerCond}
\OptFor{b}{x} = \arg \min_b \DensityHat{X \Conditional \OptFor{A}{X}, b}
,
\end{equation}
then
\begin{equation}
\label{eqn:NMLReplaceOptBySpecificLower}
\begin{split}
    &\Int{} {} {\DensityHat{X \Conditional \OptFor{A}{X}, \OptFor{b}{X}}} {X}
    \geq
    \max_{b \in \Group{B}}\Int{} {} {\DensityHat{X \Conditional \OptFor{A}{X}, b}} {X}
    .
\end{split}
\end{equation}
\end{lemma}

\newcommand{\myGlobalTransformation}[2]
{
    \pgftransformcm{-0.}{1}{0.4}{0.5}{\pgfpoint{#2cm}{#1cm}}
}

\newcommand{\gridThreeD}[5]
{
    \begin{scope}
        \myGlobalTransformation{#1}{#2};
        \draw [#3, step=0.5cm] grid (2.5, 2.5);
        \filldraw[fill=gray!40!white, draw=gray] ({#4}, {#5}) circle (0.2cm);
    \end{scope}

    \draw [dotted] (0,0, 0) -- (8,0, 0) node[midway] {$X$};
    \draw [dotted] (0,0, 0) -- (0,0, -5) node[midway] {$A$};
    \draw [dotted] (0,0, 0) -- (0,3, 0) node[midway] {$b$};
}

\tikzstyle myBG=[line width=3pt,opacity=1.0]

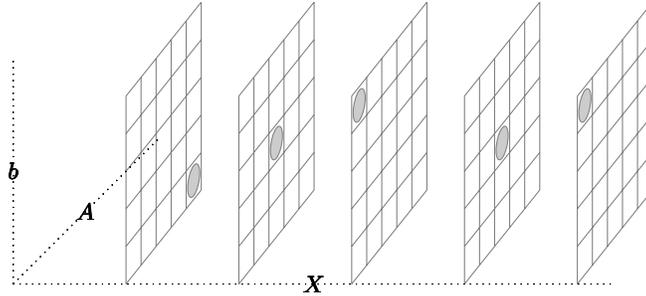
\begin{figure}[ht]
\centering
\caption{Parametric complexity using only a subset of the features. For each $X$, there are an optimal $\OptFor{A}{X}$ and $\OptFor{b}{x}$, but we wish to bound this by expressions in which for each $X$, $b$ is constant.}
\label{fig:NMLReplaceOptBySpecific}
\scalebox{\scaleFactor}{
\begin{tikzpicture}

    \gridThreeD{0}{1.5}{black!50}{0.25}{2.25};
    \gridThreeD{0}{3}{black!50}{1.25}{1.25};
    \gridThreeD{0}{4.5}{black!50}{2.25}{0.25};
    \gridThreeD{0}{6}{black!50}{1.25}{1.25};
    \gridThreeD{0}{7.5}{black!50}{2.25}{0.25};

\end{tikzpicture}
}
\end{figure}

\begin{proof}

\bigskip
For inequality (\ref{eqn:NMLReplaceOptBySpecificUpper}),
\begin{equation*}
\begin{split}
  &\Int{X} {} {\DensityHat{X \Conditional \OptFor{A}{X}, \OptFor{b}{X}}} {X}
\iftoggle{full} {
  =
}{ 
  \EqnNL{=}
} 
  \sum_{b} \Int{X:\; \OptFor{b}{X} = b} {} {\DensityHat{X \Conditional \OptFor{A}{X}, b}} {X}
  \EqnNL{ \LetteredLEq{a} }
  \sum_{b} \Int{X} {} {\DensityHat{X \Conditional \OptFor{A}{X}, b}} {X}
  ,
\end{split}
\end{equation*}
where (a) follows from the non-negativity of densities. In Figure \ref{fig:NMLReplaceOptBySpecific}, this corresponds to bounding by considering the sum of all planes, then slicing them by vertical levels.

For inequality (\ref{eqn:NMLReplaceOptBySpecificLower}), consider an arbitrary $b' \in \Group{B}$. Then
\begin{equation*}
\begin{split}
  &\Int{X} {} {\DensityHat{X \Conditional \OptFor{A}{X}, \OptFor{b}{X}}} {X}
  \iftoggle{full} {
    =
  }{ 
    \EqnNL{=}
  } 
  \sum_{b} \Int{X:\; \OptFor{b}{X} = b} {} {\DensityHat{X \Conditional \OptFor{A}{X}, b}} {X}
  \EqnNL{=}
  \Int{X:\; \OptFor{b}{X} = b'} {} {\DensityHat{X \Conditional \OptFor{A}{X}, b'}} {X}
  +
  \sum_{b \neq b'} \Int{X:\; \OptFor{b}{X} = b} {} {\DensityHat{X \Conditional \OptFor{A}{X}, b}} {X}
  \EqnNL{ \LetteredGEq{a} }
  \Int{X:\; \OptFor{b}{X} = b'} {} {\DensityHat{X \Conditional \OptFor{A}{X}, b'}} {X}
  +
  \sum_{b \neq b'} \Int{\OptFor{b}{X} = b} {} {\DensityHat{X \Conditional \OptFor{A}{X}, b'}} {X}
  \EqnNL{=}
  \Int{X} {} {\DensityHat{X \Conditional \OptFor{A}{X}, b'}} {X}
  ,
\end{split}
\end{equation*}
where (a) follows from condition (\ref{eqn:NMLReplaceOptBySpecificLowerCond}). Since this is true for an arbitrary $b'$, it is true for the maximum. In Figure \ref{fig:NMLReplaceOptBySpecific}, this corresponds to moving the disks until they are at the same horizontal level.

\end{proof}

The next section formalizes the application of the lemma to PCA NML.

\section{Reducing PCA NML to Linear Regression NML}\label{sec:TheReductions}

Let $v_{i, j}$ be the elements of the unitary matrix $V$ from Theorem \ref{thm:SVD}. By the Cauchy-Schwartz Inequality,
$\SizeOf{v_{i, j}} \leq 1$.
Let $\epsilon \lneq {1 \over m}$ be a number such that $1 \over \epsilon$ is an integer. We can quantize $v_{i, j}$ into one of ${2 \over \epsilon} + 1$ values, each distanced $\epsilon$ from each other, resulting in the matrix $V^{\epsilon}$. By considering its Neumann series, it is clear that it is invertible, so there exists some $W'$ such that $W' V^{\epsilon} = W V$.

\begin{figure}
\centering
\caption{Factor graph of known quantized loadings "PCA".}
\label{fig:KnownQuantizedLoadingPCA}
\scalebox{\scaleFactor}{
\begin{tikzpicture}

  \node[obs] (x) {$X$};
  \node[latent, above=of x, xshift=-1.2cm] (w) {$W_k$};
  \node[obs, above=of x, xshift=1.2cm] (v) {$V_k^{\epsilon}$};
  \node[obs, above=of v, xshift=-1.2cm] (k) {$k$};
  \node[latent, right=1.8cm of x] (ups) {$\Upsilon$};
  \factor[right=0.5cm of ups] {ne} {above:\NormalD} {} {}; %
  \node[const, right=0.3cm of ne] (tau) {$\tau$};
  \node[obs, right=0.5cm of v] (vd) {$\epsilon E_k$}; %

  \edge {tau} {ups} ; %
  \edge {k} {w, v, vd} ; %
  \edge {w, v, ups} {x} ; %
  \edge {vd} {v} ; %

  {\tikzset{plate caption/.append style={above=1pt of #1.south west}}
  \plate {wx} {(w)(x.south east)(w.south west)} {$n$} ;
  }
  \plate {vx} {(v)(x)(wx.north east)(x.south west)} {$m$} ;
  \plate {upsn} {(ne.north east)(ups.south west)(ups.north west)} {$nm$} ;

\end{tikzpicture}
}
\end{figure}
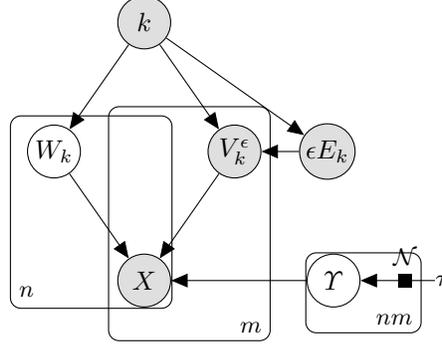

\bigSkipOnNonfull
Using Lemma \ref{lem:NMLReplaceOptBySpecific}, therefore, we can reduce the original problem to that in Figure \ref{fig:KnownQuantizedLoadingPCA}, where $V_k^{\epsilon}$ is a known matrix which is quantized version of a unitary matrix $V_k$ (specifically, $V_k^{\epsilon} = V_k + \epsilon E_k$, where $E_k$ has values each with absolute value at most $1 \over 2$). Let $\GroupVKEps$ be the set of the quantized matrices, and let $s_i^{\epsilon}(X, k)$ be the stochastic complexity of Figure \ref{fig:KnownQuantizedLoadingPCA}, where the loadings matrix is known to be the $i$th element of $\GroupVKEps$ (according to some enumeration). Then by Lemma \ref{lem:NMLReplaceOptBySpecific},
\begin{equation}\label{eqn:BoundingSBySOne}
\max_{i \in \Set{1, \ldots, \SizeOf{ \GroupVKEps }}} s_i^{\epsilon}\left(X \Given k\right)
\leq
s\left(X \Given k\right)
\leq
\sum_{i = 1}^{\SizeOf{ \GroupVKEps }} \left[ s_i^{\epsilon}\left(X \Given k\right) \right]
.
\end{equation}
Furthermore, we will see in Appendix \ref{subsec:NumQuantizedMatrices} the following lemma:
\begin{lemma}\label{lem:NumQuantizedMatrices}
\begin{equation}
\begin{split}
  \Ln{ \left|\Group{V_k^{\epsilon}}\right| }
  \lesssim
  mk \Ln{
    \left( {2 \over \epsilon} + 1 \right)
    e^{- \left( {1 - {1 + \epsilon + {\epsilon^2 \over 4}\over \sqrt{m}}\over 2} \right)}
  }
  +
  (k - 1) \Ln{ \epsilon + {m \epsilon^2 \over 4} \over \pi }
  .
\end{split}
\end{equation}
\end{lemma}

\bigskip

Let $V_k^{\epsilon}$, a known quantized loadings matrix, be the $i$th item in $\Group{V_k^{\epsilon}}$. To calculate its NML, note that Figure \ref{fig:KnownQuantizedLoadingPCA} is very similar to linear regression (whose NML is known), except that $W_k$ and $X$ are matrices instead of vectors. This can be easily reduced to linear regression, though, by considering the problem
\begin{equation*}
\begin{split}
&
\Vec{x}
=
\tilde{V_k}^{\epsilon} \Vec{w} + \Vec{\upsilon}
\EqnNL{=}
\begin{bmatrix}
  X_1^T \\
  \vdots \\
  X_n^T
\end{bmatrix}
=
\begin{bmatrix}
  V_k^{\epsilon} & \ldots & 0\\
  \vdots & \ddots & \vdots \\
  0 & \ldots & V_k^{\epsilon}
\end{bmatrix}
\begin{bmatrix}
  W_1^T \\
  \vdots \\
  W_n^T
\end{bmatrix}
+
\begin{bmatrix}
  \Upsilon_1^T \\
  \vdots \\
  \Upsilon_n^T
\end{bmatrix}
,
\end{split}
\end{equation*}
where $\Vec{x}$ and $\Vec{\upsilon}$ each have length $nm$, $\tilde{V_k}^{\epsilon}$ is $mn \times kn$, and $\Vec{w}$ has length $km$. This is the dashed part of Figure \ref{fig:LinearRegressionNML}, and has known NML (see Equation (19) in \cite{Rissanen1999MDLDenoising})
\begin{equation}\label{eqn:LinPCANML}
\begin{split}
&s_i^{\epsilon}(X, k) =
\left( nm - kn \right) \Ln{\hat{\tau}}
+ nk \Ln{ \FrobeniusSq{ \tilde{V}_k^{\epsilon} \hat{w}} }
\EqnNL{+}
(mn - kn - 1) \Ln{ mn \over mn - kn}
-
(nk + 1) \Ln{nk}
.
\end{split}
\end{equation}
However, we need the NML to be expressed in terms from the original problem.

\begin{figure}
\centering
\caption{Linear-regression factor graph.}
\label{fig:LinearRegressionNML}
\scalebox{\scaleFactor}{
\begin{tikzpicture}

  \node[obs] (x) {$\Vec{x}$};
  \node[latent, above=of x, xshift=-1.2cm] (w) {$\Vec{w}$};
  \node[latent, left=of w, xshift=-0.6cm] (origw) {$W_k$};
  \node[obs, above=of x, xshift=1.2cm] (v) {$\tilde{V_k}^{\epsilon}$};
  \node[obs, right=of v, xshift=0.3cm] (vorig) {$V_k^{\epsilon}$};
  \node[obs, above=of v, xshift=-1.2cm] (k) {$k$};
  \node[latent, right=1.8cm of x] (ups) {$\Vec{\upsilon}$};
  \factor[right=0.5cm of ups] {ne} {above:\NormalD} {} {}; %
  \node[const, right=0.3cm of ne] (tau) {$\tau$};
  \node[obs, right=0.5cm of vorig] (vd) {$\epsilon E_k$}; %

  \edge {tau} {ups} ; %
  \edge {k} {origw, vorig, vd, w, v} ; %
  \edge {w, v, ups} {x} ; %
  \edge {vorig} {v} ; %
  \edge {origw} {w} ; %
  \edge {vd} {vorig} ; %

  {\tikzset{plate caption/.append style={above=1pt of #1.south west}}
  \plate {wx} {(w)(x.south east)(w.south west)} {$n$} ;
  }
  \plate {vx} {(v)(x)(wx.north east)(x.south west)} {$m$} ;
  \plate {upsn} {(ne.north)(ups.south west)(ups.north west)} {$nm$} ;
  \plate {vorigv} {(v.south west)(v.north west)(vorig.south east)} {$n$} ;
  \draw[dashed,rounded corners=8pt] (-2,-1) -- (-2,4) -- (1.9, 4)
     -- (1.9, 0.75) -- (4.4, 0.75) -- (4.4, -1) -- cycle;

\end{tikzpicture}
}
\end{figure}

\medskip

It is well known (see \cite{Hastie01ElementsStatisticalLearning}) that
\begin{equation*}
\begin{split}
&
\Vec{\hat{w}}
=
\begin{bmatrix}
  \left(V_k^{\epsilon T} V_k^{\epsilon} \right)^{-1} V_k^{\epsilon T} X_1^T\\
  \vdots \\
  \left(V_k^{\epsilon T} V_k^{\epsilon} \right)^{-1} V_k^{\epsilon T} X_n^T
\end{bmatrix}
.
\end{split}
\end{equation*}
Furthermore, for the $j$th range,
\begin{equation*}
\begin{split}
&\hat{W}_j^T = \left(V_k^{\epsilon T} V_k^{\epsilon} \right)^{-1} V_k^{\epsilon T} X_j^T
\EqnNL{\simeq}
\left(I_k + \epsilon \left( V_k^T E + E^T V_k \right) \right)^{-1}V_k^T X_j^T
\EqnNL{\LetteredSimEq{a}}
\left(I_k - \epsilon \left( V_k^T E + E^T V_k \right) \right)V_k^T X_j^T
,
\end{split}
\end{equation*}
where (a) follows from \cite{Peterman2012MatrixHandbook} Equation (191). Therefore,
\begin{equation*}
\begin{split}
\left( V_k + \epsilon E \right)\hat{W}_j^T
\simeq
\left( I_k - \epsilon  \left({V_k^T E + E^T V_k} + E V_k^T \right) \right) X_j^T
,
\end{split}
\end{equation*}
and, finally,
\begin{equation}\label{eqn:WjtByXtX}
\begin{split}
\SizeOf{
  \Ln{ \FrobeniusSq{ V_k^{\epsilon} \hat{W}_j^T} }
  -
  \Ln{ X_j^T X_j }
}
\lesssim
 2 \epsilon
.
\end{split}
\end{equation}

\bigskip

We now prove Theorem \ref{thm:BoundPCANML}:
\begin{proof}
In equation (\ref{eqn:LinPCANML}), we replace $\hat{\tau}$
using Theorem \ref{eqn:SVDEckartYoungMirsky}, and $\tilde{V}_k^{\epsilon} \hat{w}$
using equation (\ref{eqn:WjtByXtX}).
We use the resulting expression - which is independent from $i$ (the element of $\GroupVKEps$) - in Lemma \ref{lem:NMLReplaceOptBySpecific}.
\end{proof}

\iftoggle{full} {
\section{Numerical Experiments}\label{sec:NumericalExperiments}

For numerical experiments\footnote{See \url{https://github.com/atavory/pca_nml_numerical_experiments/blob/master/numerical_experiments.ipynb} for full details.} we use the \textit{Dow-Jones Industrial Index} (DJIA), with up to 2030 days, and 30 closing prices. We transform the $i, j$-th entry, $c_{i, j}$ denoting the closing price of stock $j$ at day $i$, to $100 {c_{i, j} - c_{i - 1, j} \over c_{i - 1, j}}$, i.e., the relative closing price in percentage (see \cite{RePEc:mtp:titles:0262232197}). In the following, Orig is this matrix; Lin10 is a matrix whose first 10 columns are the original ones, and the last 20 are a random linear combination of the first 10, with $\NormalD\left(0, 0.1\right)$ noise added; Lin5 is the same, but with the last 25 generated from the first 5. By the construction, it is apparent that, at least for large enough datasets, the correct number of principal components should be 30, 10, and 5, respectively.

Consider the variance explained by the principle components for the three datasets. This is typically done via a scree plot (see \cite{Cattell1996Scree,Zhu2006AutomaticDS}), which Figure \ref{fig:NumericalScree} shows for these datasets. The horizontal axis in the plot shows the indexing of the principal components ordered by the magnitudes of eigenvalues. The vertical axis shows the variance explained by each of the components. As is typical for scree plots, the first few principal components explain much more of the variance than latter ones. In fact, there seems to be a ``bend'' in the plot for each one of the datasets, that can indicate the optimal number of components. Unfortunately, the plots for the three datasets seem to be very similar, and their ``bends'' seem to be at around the same number of components. It is not apparent to judge, by eye, what number of components should be used.

\begin{figure}[H]
\centering
\caption{Scree plots for the three datasets.}
\label{fig:NumericalScree}
\centering
\includegraphics[scale=0.75]{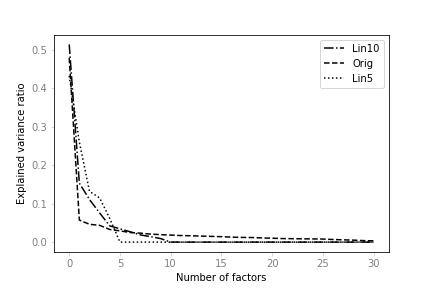}
\end{figure}

Using the Kneedle algorithm (see \cite{SatopaaAIR2011Needle}) for finding ``bends'' in plots, we get the estimated optimal number of components, as a function of the dataset length, in Figure \ref{fig:NumericalScreeKnee}. This method is known for its tendency to find a lower number of components than the true one (see \cite{Wayne2010Clinical}), as is indeed the case here.

\begin{figure}[ht]
\centering
\caption{Optimal components using the knee method.}
\label{fig:NumericalScreeKnee}
\centering
\includegraphics[scale=0.75]{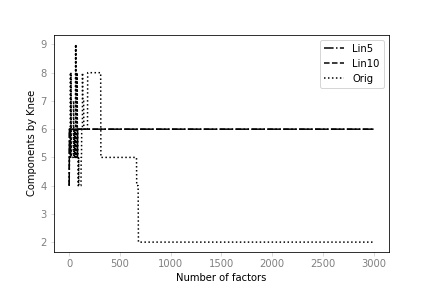}
\end{figure}

The Kaiser method (see \cite{Jolliffe1986PCA}) takes components whose eigenvalues are at least one. Figure \ref{fig:NumericalScreeKaiser} shows the estimated optimal number of components, as a function of the dataset length, using this method. While this method does better, it also underestimates the number of components. It is also interesting to note that the results are not monotone in the length of the datasets.

\begin{figure}[ht]
\centering
\caption{Optimal components using the Kaiser method.}
\label{fig:NumericalScreeKaiser}
\centering
\includegraphics[scale=0.75]{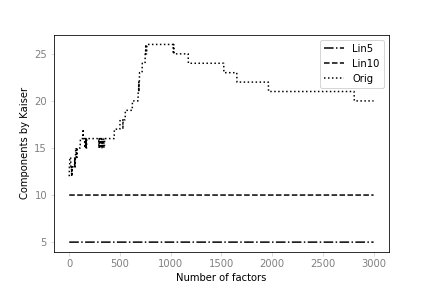}
\end{figure}

Finally, Figure \ref{fig:NumericalOptK} shows the upper and lower bounds for the optimal number of components as a function of dataset length, using the NML technique from this paper. Note that we don't have an analytical expression for the NML of PCA, but rather bounds for it. Figure \ref{fig:NumericalOptKRatio} shows the ratio of the bounds as a function of the dataset length.

\begin{figure}[ht]
\centering
\caption{Lower and upper bounds for the optimal $\hat{k}$, for the three datasets.}
\centering
\includegraphics[scale=0.75]{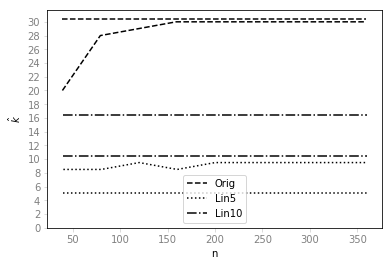}
\label{fig:NumericalOptK}
\end{figure}

\begin{figure}[ht]
\centering
\caption{Relative change between the upper and lower bounds of the NML, compared to the NML, for the three datasets.}
\label{fig:NumericalRelative}
\centering
\includegraphics[scale=0.75]{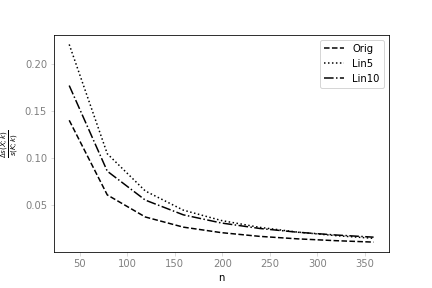}
\label{fig:NumericalOptKRatio}
\end{figure}

}{ 
} 

\section{Conclusions and Future Work}\label{sec:Conclusions}

In this work we saw an NML-calculation technique based on reducing a problem through eliminating the optimization of some of its original dimensions. We saw how to use this to bound the NML of PCA. The technique is simple and general, and can be used to reduce problems in other domains, where simpler versions of the problem have a closed-form NML. Unfortunately, there are also several types of simple problems with no closed-form NML. For these cases, an MCMC evaluation of the parametric complexity (the denominator of Equation (\ref{eqn:NMLDef})), could be a good numeric approximation. Developing an efficient algorithm for this, is a topic for further research.

\appendix

\section{Appendix}\label{sec:Appendix}

\subsection{Number of Quantized Unitary Matrices}\label{subsec:NumQuantizedMatrices}

We prove here Lemma \ref{lem:NumQuantizedMatrices}. Let $\Vec{v_i},
\Vec{v_j}$ be two columns of a
unitary matrix (perhaps the one), and $\Vec{v_i^{\epsilon}}, \Vec{v_j^{\epsilon}}$
be their quantized
counterparts. Simple arithmetic shows that
\begin{equation}\label{eqn:ItIsAlmostOne}
\left|
\Vec{v_i^{\epsilon}} \cdot \Vec{v_j^{\epsilon}} - \Vec{v_i} \cdot \Vec{v_j}
\right|
\leq
\epsilon
+
{m \epsilon^2 \over 4}
.
\end{equation}
\bigSkipOnNonfull
We will see that
\begin{equation}
\label{eqn:NM}
\begin{split}
&\Probability{
  \left| \Vec{v_i^{\epsilon}} \cdot \Vec{v_i^{\epsilon}} \right|
  \leq
  1 + \epsilon + {\epsilon^2 \over 4}
}
\leq
e^{-mk {1 - {{1 + \epsilon + {\epsilon^2 \over 4}} \over \sqrt{m}} \over 2}}
,
\\
&\Probability{
  \left| \Vec{v_{i + 1}^{\epsilon}} \cdot \Vec{v_i^{\epsilon}} \right|
  \leq
  \epsilon + {\epsilon^2 \over 4}
  |\;
  \forall_{j} \Vec{v_j} \Vec{v_j^{\epsilon}} \leq 1 + \epsilon + {\epsilon^2 \over 4}
}
\leq
\left(
{\epsilon + {m \epsilon^2 \over 4}} \over \pi
\right)^{k - 1}
.
\end{split}
\end{equation}

\bigSkipOnNonfull

For the first part of inequality (\ref{eqn:NM}),
\begin{equation}
\begin{split}
&\Probability{\sum_{k = 1}^m\left[ v_{i, k}^2\right] \leq 1 + \epsilon + {m \epsilon^2 \over 4}}
\iftoggle{full} {
  \leq
}{ 
  \EqnNL{\leq}
} 
\Probability{\SizeOf{ \Set{k \Conditional v_{i, k}^2 \geq x } } \leq {1 + \epsilon + {m \epsilon^2 \over 4} \over x}}
\EqnNL{\LetteredLEq{a}}
e^{ -m \Kullback{1 \over mx}{ \sqrt{1 - x} } }
,
\end{split}
\end{equation}
where (a) follow from the Chernoff bound (see \cite{MitzenmacherUpfal2010ProbabilityComputing}, Chapter
5). Using the well-known bound (see \cite{Cover2006Elements}, \cite{MitzenmacherUpfal2010ProbabilityComputing}, Chapter 5),
\begin{equation*}
\Kullback{x}{y} \geq {(x - y)^2 \over 2y}, \; (x \leq y)
,
\end{equation*}
and so
\begin{equation}
\begin{split}
&\Kullback{1 + \epsilon + {m \epsilon^2 \over 4} \over mx}{ \sqrt{1 - x} }
\geq
{ \left( {1 + \epsilon + {m \epsilon^2 \over 4} \over mx} - \sqrt{1 - x} \right)^2 \over 2 \sqrt{1 - x} }
.
\end{split}
\end{equation}
Setting $x = {1 \over \sqrt{m}}$, we have
\begin{equation}
\begin{split}
&\Kullback{1 + \epsilon + {m \epsilon^2 \over 4} \over \sqrt{m}}{ \sqrt{1 - {1 \over \sqrt{m}}} }
\iftoggle{full} {
  \geq
}{ 
  \EqnNL{\geq}
} 
{ \left( {1 + \epsilon + {m \epsilon^2 \over 4} \over \sqrt{m}} - \sqrt{1 - {1 \over \sqrt{m}}} \right)^2 \over 2 \sqrt{1 - {1 \over \sqrt{m}}} }
\EqnNL{\LetteredSimEq{a}}
{ \left( 1 - {3 + \epsilon + {\epsilon^2 \over 4} \over 2 \sqrt{m}} \right)^2 \over 2 \left(1 - {2 \over \sqrt{m}} \right)}
\iftoggle{full} {
  \LetteredSimEq{b}
}{ 
  \EqnNL{\LetteredSimEq{b}}
} 
{1 - {1 + \epsilon + {\epsilon^2 \over 4}\over \sqrt{m}}\over 2},
\end{split}
\end{equation}
where (a) and (b) follow from the Taylor expansion of $(1 + x)^{\alpha}$.

\bigSkipOnNonfull

For the second part of Inequality (\ref{eqn:NM}), applying equation (\ref{eqn:ItIsAlmostOne}) twice on the left side, and once on the right side, we have
\begin{equation}
\Vec{v_i^{\epsilon}} \cdot \Vec{v_j^{\epsilon}} = \LTwo{\Vec{v_i^{\epsilon}}}\LTwo{\Vec{v_j^{\epsilon}}} \Cos{\alpha_{i, k}^{\epsilon}}
,
\end{equation}
and so
\begin{equation}
{\pi \over 2} - \alpha_{i, k}^{\epsilon}
\LetteredSimEq{a}
\Sin{{\pi \over 2} - \alpha_{i, k}^{\epsilon}}
=
\Cos{\alpha_{i, k}^{\epsilon}}
\leq
\SizeOf{
  {\epsilon + {m \epsilon^2 \over 4} \over 1 - \epsilon - {m \epsilon^2 \over 4}}
}
,
\end{equation}
with $\alpha_{i, k}^{\epsilon}$ the angle between the vectors, and where (a) follows from the Taylor series of $\Sin{x}$.
Approximating
$\alpha_{i, k}^{\epsilon} \sim \UniformD\left(0, 2 \pi\right)$, we get that the probability is approximately
that in the second part of Inequality (\ref{eqn:NM}).

\bibliographystyle{plain}
\bibliography{bibli}
\end{document}